\documentclass[11pt, a4paper, logo, twocolumn, copyright]{googledeepmind}

\usepackage[authoryear, sort&compress, round]{natbib}
\title{ConvNets Match Vision Transformers at Scale}

\correspondingauthor{slsmith@google.com}

\reportnumber{} % Leave blank if n/a

\author[1]{Samuel L Smith}
\author[1]{Andrew Brock}
\author[1]{Leonard Berrada}
\author[1]{Soham De}
\affil[1]{Google DeepMind}
\begin{abstract}
Many researchers believe that ConvNets perform well on small or moderately sized datasets, but are not competitive with Vision Transformers when given access to datasets on the web-scale. We challenge this belief by evaluating a performant ConvNet architecture pre-trained on JFT-4B, a large labelled dataset of images often used for training foundation models. We consider pre-training compute budgets between 0.4k and 110k TPU-v4 core compute hours, and train a series of networks of increasing depth and width from the NFNet model family. We observe a log-log scaling law between held out loss and compute budget. After fine-tuning on ImageNet, NFNets match the reported performance of Vision Transformers with comparable compute budgets. Our strongest fine-tuned model achieves a Top-1 accuracy of 90.4$\%$.

\end{abstract}

\begin{document}

\maketitle

\section*{Introduction}

Convolutional Neural Networks (ConvNets) were responsible for many of the early successes of deep learning. Deep ConvNets were first deployed commercially over 20 years ago \citep{lecun1998gradient}, while the success of AlexNet on the ImageNet challenge in 2012 re-ignited widespread interest in the field \citep{krizhevsky2017imagenet}. For almost a decade ConvNets (typically ResNets \citep{he2016deep, he2016identity}) dominated computer vision benchmarks. However in recent years they have increasingly been replaced by Vision Transformers (ViTs) \citep{dosovitskiy2020image}. 

Simultaneously, the computer vision community has shifted from primarily evaluating the performance of randomly initialized networks on specific datasets like ImageNet, to evaluating the performance of networks pre-trained on large general purpose datasets collected from the web. This raises an important question; do Vision Transformers outperform ConvNet architectures pre-trained with similar computational budgets?

Although most researchers in the community believe Vision Transformers show better scaling properties than ConvNets, there is surprisingly little evidence to support this claim. Many papers studying ViTs compare to weak ConvNet baselines (typically the original ResNet architecture \citep{he2016deep}). Additionally, the strongest ViT models have been pre-trained using large compute budgets beyond 500k TPU-v3 core hours \citep{zhai2022scaling}, which significantly exceeds the compute used to pre-train ConvNets.

\begin{figure*}[t]
	\centering
	\vskip - 3mm
    \includegraphics[width=0.91\linewidth]{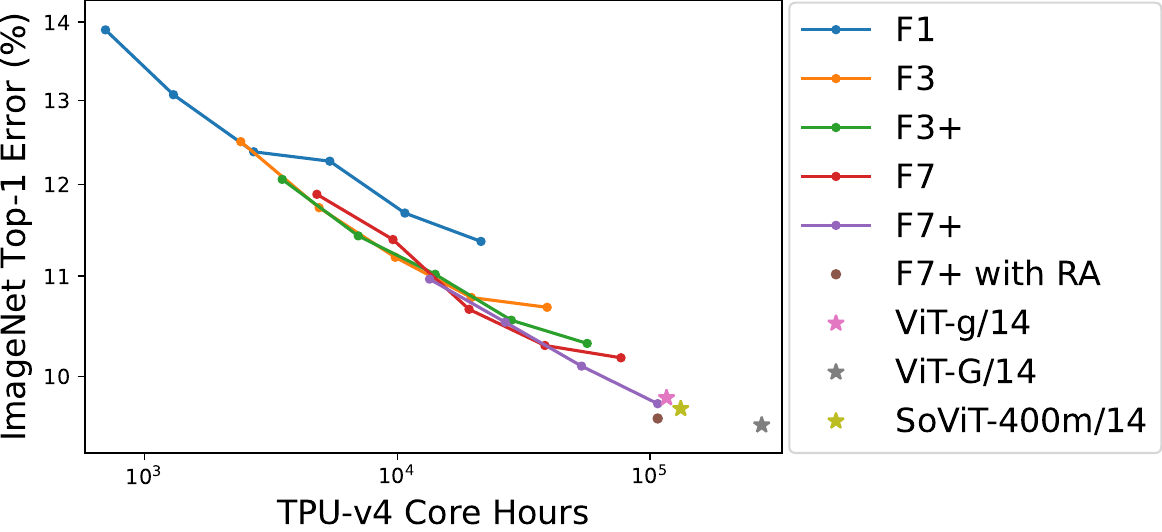}
    \vskip - 1mm
	\caption{ImageNet Top-1 error, after fine-tuning pre-trained NFNet models for 50 epochs. Both axes are log-scaled. Performance improves consistently as the compute used during pre-training increases. Our largest model (F7+) achieves comparable performance to that reported for pre-trained ViTs with a similar compute budget \citep{alabdulmohsin2023getting, zhai2022scaling}. The performance of this model improved further when fine-tuned with repeated augmentation (RA) \citep{hoffer2019augment}.}
	\vskip -2mm
	\label{fig:imagenet}
\end{figure*}

We evaluate the scaling properties of the NFNet model family \citep{brock2021high}, a pure convolutional architecture published concurrently with the first ViT papers, and the last ConvNet to set a new SOTA on ImageNet. We do not make any changes to the model architecture or the training procedure (beyond tuning simple hyper-parameters such as the learning rate or epoch budget). We consider compute budgets up to a maximum of 110k TPU-v4 core hours,\footnote{TPU-v4 cores have roughly double the theoretical flops of TPU-v3 cores, however both cores have similar memory.} and pre-train on the JFT-4B dataset which contains roughly 4 billion labelled images from 30k classes \citep{sun2017revisiting}. We observe a log-log scaling law between validation loss and the compute budget used to pre-train the model. After fine-tuning on ImageNet, our networks match the performance of pre-trained ViTs with comparable compute budgets \citep{zhai2022scaling, alabdulmohsin2023getting}, as shown in Figure \ref{fig:imagenet}.

\section*{Pre-trained NFNets obey scaling laws}
\begin{figure}[t]
	\centering
	\vskip - 3mm
    \includegraphics[width=0.95\linewidth]{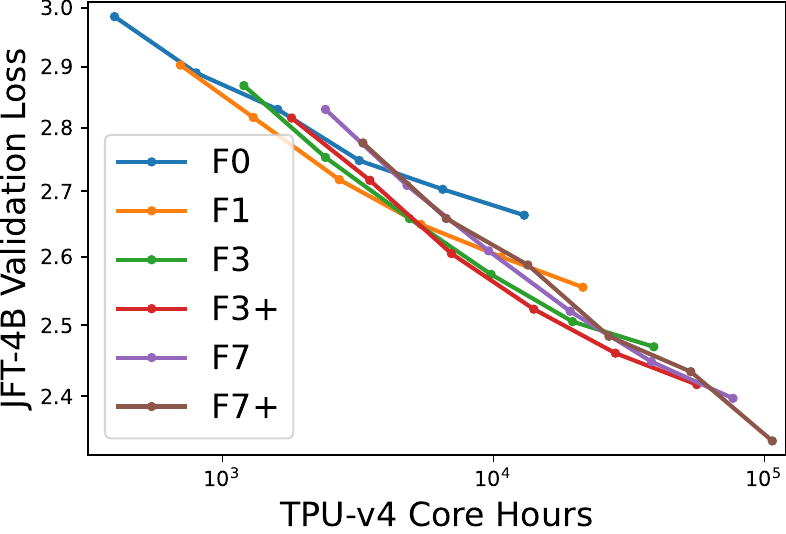}
    \vskip - 1mm
	\caption{Held out loss of NFNets on JFT-4B, plotted against the compute used during training. Both axes are log-scaled, and each curve denotes a different model trained for a range of epoch budgets. We observe a linear trend, matching the scaling laws observed for language modelling.}
	\vskip -2mm
	\label{fig:main}
\end{figure}

We train a range of NFNet models of varying depth and width on JFT-4B. Each model is trained for a range of epoch budgets between 0.25 and 8, using a cosine decay learning rate schedule. The base learning rate is tuned separately for each epoch budget on a small logarithmic grid. In Figure \ref{fig:main}, we provide the validation loss at the end of training on a held out set of 130k images, plotted against the compute budget required to train each model\footnote{We estimate the compute required to train each model by eye from the typical steps per second achieved by each model during training (when not pre-empted).}. We note that F7 has the same width as F3, but is double the depth. Similarly F3 is double the depth of F1, and F1 is double the depth of F0. F3+ and F7+ have the same depths as F3 and F7 but larger width. We train using SGD with Momentum and Adaptive Gradient Clipping (AGC) at batch size $4096$, and we use an image resolution of $224\times224$ during training and $256\times 256$ at evaluation. For additional details describing the NFNet architecture and training pipeline we refer the reader to the original paper \citep{brock2021high}, including the pre-training framework for JFT described in Section 6.2. Note that we removed near-duplicates of images in the training and validation sets of ImageNet from JFT-4B before training \citep{kolesnikov2020big}.

Figure \ref{fig:main} shows a clear linear trend, consistent with a log-log scaling law between validation loss and pre-training compute. This matches the log-log scaling laws previously observed when performing language modelling with transformers \citep{brown2020language, hoffmann2022training}. 

The optimal model size and the optimal epoch budget (which achieve the lowest validation loss) both increase in size as the compute budget increases. We found that a reliable rule of thumb is to scale the model size and the number of training epochs at the same rate, as previously observed for language modelling by \citet{hoffmann2022training}. We note that the optimal epoch budget was greater than 1 for overall compute budgets greater than roughly 5k TPU-v4 core hours.

\begin{figure}[t]
	\centering
	\vskip - 3mm
    \includegraphics[width=0.95\linewidth]{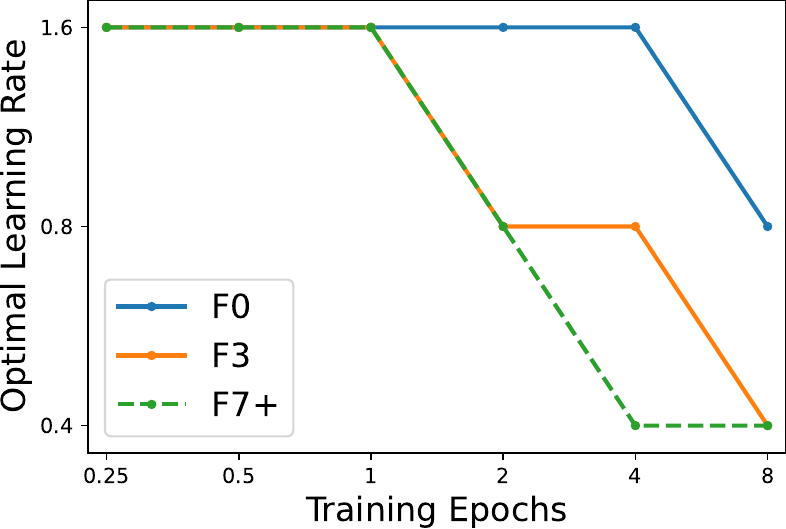}
    \vskip - 1mm
	\caption{The optimal learning rate behaves predictably and is easy to tune. All models show similar optimal learning rates $\alpha \sim 1.6$ when the epoch budget is small. The learning rate falls slowly as model size and epoch budget increases.}
	\vskip -2mm
	\label{fig:lrs}
\end{figure}

In Figure \ref{fig:lrs} we plot the observed optimal learning rate (which minimizes validation loss), for 3 of our models, across a range of epoch budgets.\footnote{The optimal learning rate showed very similar trends for all models. We select 3 models here for visual clarity.} Note that we tuned the learning rate on a logarithmic grid spaced by factors of 2. We find that all models in the NFNet family show a similar optimal learning rate $\alpha \approx 1.6$ for small epoch budgets. However the optimal learning rate falls as the epoch budget rises, and for large models the optimal learning rate falls more quickly. In practice one can efficiently tune the learning rate within 2 trials by assuming that the optimal learning rate falls slowly but monotonically as both the model size and the epoch budget increases.

Finally, we note that some pre-trained models in Figure \ref{fig:main} perform less well than expected. For example, the curve for NFNet-F7+ models at different pre-training budgets is not smooth. We believe this arises because our data loading pipeline did not guarantee that each training example would be sampled once per epoch if the training run was pre-empted/restarted, potentially causing some training examples to be under-sampled if a training run was restarted multiple times.

\section*{Fine-tuned NFNets are competitive with Vision Transformers on ImageNet}

In Figure \ref{fig:imagenet}, we fine-tune our pre-trained NFNets on ImageNet, and plot the Top-1 error against the compute used during pre-training. We fine-tune each model for 50 epochs using sharpness aware minimization (SAM) \citep{foret2020sharpness} with stochastic depth and dropout. We train at resolution $384\times384$ and evaluate at $480\times480$.

The ImageNet Top-1 accuracy consistently improves as the compute budget increases. Our most expensive pre-trained model, an NFNet-F7+ pre-trained for 8 epochs, achieves an ImageNet Top-1 accuracy of 90.3$\%$ while requiring roughly 110k TPU-v4 core hours to pre-train and 1.6k TPU-v4 core hours to fine-tune. Furthermore, we achieve 90.4$\%$ Top-1 accuracy if we additionally introduce repeated augmentation during fine-tuning \citep{hoffer2019augment, fort2021drawing} with augmentation multiplicity 4.\footnote{When using repeated augmentation, we reduce the number of passes through the data such that the total computational cost of fine-tuning is constant.} For comparison, the best reported Top-1 accuracy of an NFNet on ImageNet without extra data is 86.8$\%$ \citep{fort2021drawing}, achieved by an NFNet-F5 with repeated augmentation. This demonstrates that NFNets benefit substantially from large scale pre-training.

Despite the substantial differences between the two model architectures, the performance of pre-trained NFNets at scale is remarkably similar to the performance of pre-trained Vision Transformers. For example, \cite{zhai2022scaling} achieve 90.2$\%$ Top-1 on ImageNet with a ViT-g/14, after pre-training on JFT-3B for 210k TPU-v3 core hours, and 90.45$\%$ with a ViT-G/14 after pre-training on JFT-3B for over 500k TPU-v3 core hours. In a recent work, \cite{alabdulmohsin2023getting} optimize the ViT architecture and achieve 90.3$\%$ Top-1 with a SoViT-400m/14 after pre-training on JFT-3B for 230k TPU-v3 hours. 

We evaluated the pre-training speed for these models on TPU-v4 (using the original authors' codebase), and estimate that ViT-g/14 would take 120k TPU-v4 core hours to pre-train, while ViT-G/14 would take 280k TPU-v4 core hours and SoViT-400m/14 would take 130k TPU-v4 core hours. We use these estimates to compare the pre-training efficiency of ViTs and NFNets in Figure \ref{fig:imagenet}. We note however that NFNets were optimized for TPU-v4, and perform less well when evaluated on other devices. For example, we estimate that NFNet-F7+ would require 250 TPU-v3 core hours to pre-train for 8 epochs in our codebase.

Finally, we note that the pre-trained checkpoints achieving the lowest validation loss on JFT-4B did not always achieve the highest Top-1 accuracy on ImageNet after fine-tuning. In particular, we found that, under a fixed pre-training compute budget, the fine-tuning regime consistently favoured slightly larger models and slightly smaller epoch budgets. Intuitively, larger models have more capacity and are therefore better able to adapt to the new task. In some cases, slightly larger learning rates (during pre-training) also achieved better performance after fine-tuning.

%Finally, we note that we consistently observed that the models which achieved the lowest validation loss on JFT-4B did not also achieve the highest top-1 accuracy on ImageNet. Instead, we find that fine-tuning on ImageNet consistently favoured slightly smaller models trained for fewer epochs.

\section*{Discussion}

Our work reinforces the bitter lesson. The most important factors determining the performance of a sensibly designed model are the compute and data available for training\footnote{By sensibly designed, we mean models that are sufficiently expressive and have stable gradient propagation.} \citep{tolstikhin2021mlp}. Although the success of ViTs in computer vision is extremely impressive, in our view there is no strong evidence to suggest that pre-trained ViTs outperform pre-trained ConvNets when evaluated fairly. We note however that ViTs may have practical advantages in specific contexts, such as the ability to use similar model components across multiple modalities \citep{fuyu-8b}.

%We note that promising results have been obtained using Conv-Attention hybrids \citep{dai2021coatnet}, and we anticipate that these models may outperform both ViTs and ConvNets in the long run.

%Our work reinforces the famous bitter lesson. The most important factors determining the performance of a sensibly designed model is the compute and data available for training. %We demonstrate that a SOTA computer vision model from 2021, when scaled up, is competitive with the SOTA in 2023. This is particularly striking, since we make no changes to either the NFNet architecture or the training or fine-tuning process (other than using a larger pre-training dataset).

\section*{Acknowledgements}

We thank Lucas Beyer and Olivier Henaff for feedback on an earlier draft of this note. We also thank Lucas Beyer for providing training speed estimates for ViT models on TPU-v4 devices.

\bibliography{main}

\begin{thebibliography}{17}
\providecommand{\natexlab}[1]{#1}
\providecommand{\url}[1]{\texttt{#1}}
\expandafter\ifx\csname urlstyle\endcsname\relax
  \providecommand{\doi}[1]{doi: #1}\else
  \providecommand{\doi}{doi: \begingroup \urlstyle{rm}\Url}\fi

\bibitem[Alabdulmohsin et~al.(2023)Alabdulmohsin, Zhai, Kolesnikov, and
  Beyer]{alabdulmohsin2023getting}
I.~Alabdulmohsin, X.~Zhai, A.~Kolesnikov, and L.~Beyer.
\newblock Getting vit in shape: Scaling laws for compute-optimal model design.
\newblock \emph{arXiv preprint arXiv:2305.13035}, 2023.

\bibitem[Bavishi et~al.(2023)Bavishi, Elsen, Hawthorne, Nye, Odena, Somani, and
  Ta\c{s}\i{}rlar]{fuyu-8b}
R.~Bavishi, E.~Elsen, C.~Hawthorne, M.~Nye, A.~Odena, A.~Somani, and
  S.~Ta\c{s}\i{}rlar.
\newblock Introducing our multimodal models, 2023.
\newblock URL \url{https://www.adept.ai/blog/fuyu-8b}.

\bibitem[Brock et~al.(2021)Brock, De, Smith, and Simonyan]{brock2021high}
A.~Brock, S.~De, S.~L. Smith, and K.~Simonyan.
\newblock High-performance large-scale image recognition without normalization.
\newblock In \emph{International Conference on Machine Learning}, pages
  1059--1071. PMLR, 2021.

\bibitem[Brown et~al.(2020)Brown, Mann, Ryder, Subbiah, Kaplan, Dhariwal,
  Neelakantan, Shyam, Sastry, Askell, et~al.]{brown2020language}
T.~Brown, B.~Mann, N.~Ryder, M.~Subbiah, J.~D. Kaplan, P.~Dhariwal,
  A.~Neelakantan, P.~Shyam, G.~Sastry, A.~Askell, et~al.
\newblock Language models are few-shot learners.
\newblock \emph{Advances in neural information processing systems},
  33:\penalty0 1877--1901, 2020.

\bibitem[Dosovitskiy et~al.(2020)Dosovitskiy, Beyer, Kolesnikov, Weissenborn,
  Zhai, Unterthiner, Dehghani, Minderer, Heigold, Gelly,
  et~al.]{dosovitskiy2020image}
A.~Dosovitskiy, L.~Beyer, A.~Kolesnikov, D.~Weissenborn, X.~Zhai,
  T.~Unterthiner, M.~Dehghani, M.~Minderer, G.~Heigold, S.~Gelly, et~al.
\newblock An image is worth 16x16 words: Transformers for image recognition at
  scale.
\newblock \emph{arXiv preprint arXiv:2010.11929}, 2020.

\bibitem[Foret et~al.(2020)Foret, Kleiner, Mobahi, and
  Neyshabur]{foret2020sharpness}
P.~Foret, A.~Kleiner, H.~Mobahi, and B.~Neyshabur.
\newblock Sharpness-aware minimization for efficiently improving
  generalization.
\newblock \emph{arXiv preprint arXiv:2010.01412}, 2020.

\bibitem[Fort et~al.(2021)Fort, Brock, Pascanu, De, and Smith]{fort2021drawing}
S.~Fort, A.~Brock, R.~Pascanu, S.~De, and S.~L. Smith.
\newblock Drawing multiple augmentation samples per image during training
  efficiently decreases test error.
\newblock \emph{arXiv preprint arXiv:2105.13343}, 2021.

\bibitem[He et~al.(2016{\natexlab{a}})He, Zhang, Ren, and Sun]{he2016deep}
K.~He, X.~Zhang, S.~Ren, and J.~Sun.
\newblock Deep residual learning for image recognition.
\newblock In \emph{Proceedings of the IEEE conference on computer vision and
  pattern recognition}, pages 770--778, 2016{\natexlab{a}}.

\bibitem[He et~al.(2016{\natexlab{b}})He, Zhang, Ren, and Sun]{he2016identity}
K.~He, X.~Zhang, S.~Ren, and J.~Sun.
\newblock Identity mappings in deep residual networks.
\newblock In \emph{European conference on computer vision}, pages 630--645.
  Springer, 2016{\natexlab{b}}.

\bibitem[Hoffer et~al.(2019)Hoffer, Ben-Nun, Hubara, Giladi, Hoefler, and
  Soudry]{hoffer2019augment}
E.~Hoffer, T.~Ben-Nun, I.~Hubara, N.~Giladi, T.~Hoefler, and D.~Soudry.
\newblock Augment your batch: better training with larger batches.
\newblock \emph{arXiv preprint arXiv:1901.09335}, 2019.

\bibitem[Hoffmann et~al.(2022)Hoffmann, Borgeaud, Mensch, Buchatskaya, Cai,
  Rutherford, Casas, Hendricks, Welbl, Clark, et~al.]{hoffmann2022training}
J.~Hoffmann, S.~Borgeaud, A.~Mensch, E.~Buchatskaya, T.~Cai, E.~Rutherford,
  D.~d.~L. Casas, L.~A. Hendricks, J.~Welbl, A.~Clark, et~al.
\newblock Training compute-optimal large language models.
\newblock \emph{arXiv preprint arXiv:2203.15556}, 2022.

\bibitem[Kolesnikov et~al.(2020)Kolesnikov, Beyer, Zhai, Puigcerver, Yung,
  Gelly, and Houlsby]{kolesnikov2020big}
A.~Kolesnikov, L.~Beyer, X.~Zhai, J.~Puigcerver, J.~Yung, S.~Gelly, and
  N.~Houlsby.
\newblock Big transfer (bit): General visual representation learning.
\newblock In \emph{Computer Vision--ECCV 2020: 16th European Conference,
  Glasgow, UK, August 23--28, 2020, Proceedings, Part V 16}, pages 491--507.
  Springer, 2020.

\bibitem[Krizhevsky et~al.(2017)Krizhevsky, Sutskever, and
  Hinton]{krizhevsky2017imagenet}
A.~Krizhevsky, I.~Sutskever, and G.~E. Hinton.
\newblock Imagenet classification with deep convolutional neural networks.
\newblock \emph{Communications of the ACM}, 60\penalty0 (6):\penalty0 84--90,
  2017.

\bibitem[LeCun et~al.(1998)LeCun, Bottou, Bengio, and
  Haffner]{lecun1998gradient}
Y.~LeCun, L.~Bottou, Y.~Bengio, and P.~Haffner.
\newblock Gradient-based learning applied to document recognition.
\newblock \emph{Proceedings of the IEEE}, 86\penalty0 (11):\penalty0
  2278--2324, 1998.

\bibitem[Sun et~al.(2017)Sun, Shrivastava, Singh, and Gupta]{sun2017revisiting}
C.~Sun, A.~Shrivastava, S.~Singh, and A.~Gupta.
\newblock Revisiting unreasonable effectiveness of data in deep learning era.
\newblock In \emph{Proceedings of the IEEE international conference on computer
  vision}, pages 843--852, 2017.

\bibitem[Tolstikhin et~al.(2021)Tolstikhin, Houlsby, Kolesnikov, Beyer, Zhai,
  Unterthiner, Yung, Steiner, Keysers, Uszkoreit, et~al.]{tolstikhin2021mlp}
I.~O. Tolstikhin, N.~Houlsby, A.~Kolesnikov, L.~Beyer, X.~Zhai, T.~Unterthiner,
  J.~Yung, A.~Steiner, D.~Keysers, J.~Uszkoreit, et~al.
\newblock Mlp-mixer: An all-mlp architecture for vision.
\newblock \emph{Advances in neural information processing systems},
  34:\penalty0 24261--24272, 2021.

\bibitem[Zhai et~al.(2022)Zhai, Kolesnikov, Houlsby, and
  Beyer]{zhai2022scaling}
X.~Zhai, A.~Kolesnikov, N.~Houlsby, and L.~Beyer.
\newblock Scaling vision transformers.
\newblock In \emph{Proceedings of the IEEE/CVF Conference on Computer Vision
  and Pattern Recognition}, pages 12104--12113, 2022.

\end{thebibliography}

\end{document}